\documentclass[letterpaper,conference]{IEEEtran} 
\usepackage{mathptmx} %

\newcommand{\ignore}[1]{}
\usepackage{fancyhdr}
\usepackage[normalem]{ulem}
\usepackage[hyphens]{url}

\usepackage{amsmath}
\usepackage{graphicx}
\usepackage{xcolor}
\usepackage{subcaption}
\usepackage{xspace}
\usepackage{bbold}
\usepackage{pbox}
\usepackage[rightcaption]{sidecap}
\usepackage{flushend}
\sidecaptionvpos{figure}{c}

\definecolor{Awesome}{rgb}{1.0, 0.08, 0.58}

\newcommand{\ZFL}{\textit{Cnvlutin}\xspace}
\newcommand{\ZF}{\textit{CNV}\xspace}
\newcommand{\ZFLn}{\textit{Cnvlutin\textsuperscript{2}}\xspace}

\newcommand{\ZFNAf}{ZFNAf\xspace}
\newcommand{\BASE}{DaDianNao\xspace}

 \fancypagestyle{firstpage}{
   \fancyhf{}
 \setlength{\headheight}{50pt}

}  

\begin{document}

\title{Cnvlutin\textsuperscript{2}: Ineffectual-Activation-and-Weight-Free Deep Neural Network Computing}
\author{Patrick Judd \quad Alberto Delmas Lascorz \quad Sayeh Sharify \quad Andreas Moshovos\\ 
\\
  University of Toronto}

\author{
Patrick Judd,
Alberto Delmas,
Sayeh Sharify
\& Andreas Moshovos\\
Department of Electrical and Computer Engineering, University of Toronto\\ 
\texttt{\{juddpatr, delmasl1, sayeh, moshovos\}@ece.utoronto.ca} \\ 
}

\maketitle
\thispagestyle{firstpage}

\pagestyle{plain}

\begin{abstract}
We discuss several modifications and extensions over the previous proposed \textit{Cnvlutin}
(\ZF) \ accelerator for convolutional and fully-connected layers of Deep Learning Network. 
We first describe different encodings of the activations that are deemed ineffectual.  The encodings have different memory overhead and energy characteristics.  We propose using a level of indirection when accessing activations from memory to reduce their memory footprint by storing only the effectual activations. We also present a modified organization that detects the activations that are deemed as ineffectual while fetching them from memory. This is different than the original design that instead detected them at the output of the preceding layer. Finally, we 
present an extended \ZF that can also skip ineffectual weights.

\end{abstract}

\section{Introduction}

Albericio \textit{et al.} proposed the \textit{Cnvlutinn } (\ZF) accelerator~\cite{cnvlutin} which exploits ineffectual activation values to improve performance and energy efficiency over conventional hardware that processes all activations regardless of their value content. \ZF benefits fully-connected and convolutional layers and was evaluated for image classification convolutional neural networks (CNNs). \ZF detects the ineffectual activations at runtime and thus does not require purpose-trained Neural Networks (NNs). This work extends the work of Albericio \textit{et al.}~\cite{cnvlutin} by discussing design alternatives and extensions over the originally proposed design.

Specifically, this work presents the following: 1) Different ways of encoding which activation values are ineffectual. The proposed encodings reduce memory storage and energy overhead compared to the original proposal. 2) An extension where the detection of ineffectual activations is done while fetching them from memory. This design presents no memory  storage and energy overhead as it obviates the needs to store explicit information about which activations are ineffectual. 3) An extension where only the effectual activations are stored in memory which can reduce memory footprint. Finally, 4) this work presents an extension to \ZF that eliminates computations involving not only ineffectual activations but also ineffectual weights. All aforementioned alternatives and extensions do not modify the core execution engines of \ZF. Instead, they require changes only to the \textit{dispatcher} and the \textit{reducer. }In the original \ZF\ the dispatcher  fetches activations and distributes them to the execution units, while the Reducer writes the output activations to memory.  
\section{Background}
\subsection{DaDianNao}

\ZFL\ was shown to improve performance and energy over  the data-parallel  DaDianNao (DaDN)\ accelerator~\cite{DaDiannao}. This work uses the same baseline for consistency. DaDN processes all activations regardless of their values. DaDN is a massively data-parallel architecture. Every cycle, it processes 16 activation values, and weights from up to 256 filters. Specifically, for each filter, DaDN multiplies the 16 activation values with 16 weights and accumulates the result into a partial output activation. This process repeats until all activation values necessary have been processed for each desired output activation.  

\begin{figure*}[htb!]
        \centering
        \includegraphics[width=\textwidth]{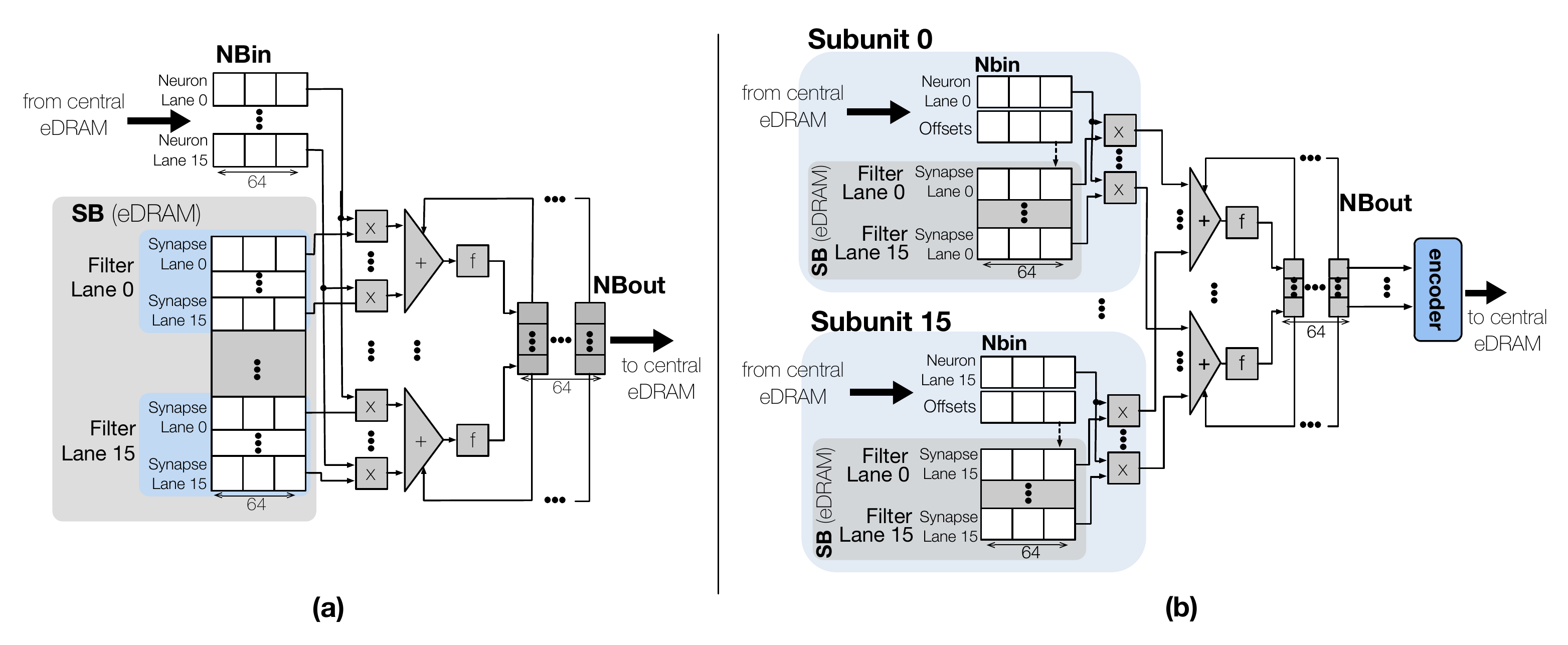}
   
\caption{Compute Units. \textbf{a)} DaDianNao NFU. \textbf{b)} \ZF unit. From Albericio \textit{et al.}~\cite{cnvlutin}}
\label{fig:nodes-overview}
\end{figure*}

Each \BASE chip, or node, contains 16 \textit{Neural Functional Units (NFUs)}, or simply \textit{units}. Figure~\ref{fig:nodes-overview}(a) shows one such unit. Each cycle the unit processes 16 input activations, 256 weights from 16 filters, and produces 16 partial output activations. In detail, the unit has 16 neuron lanes,\footnote{The original DaDianNao publication uses the terms neuron and synapse instead of the more commonly used terms activations and weights. We maintain their terminology for the phyical structures} 16 filter lanes each with 16 synapse lanes (256 in total), and produces 16 partial sums for 16 output activations. The unit's SB has 256 lanes ($16\times 16$) feeding the 256 synapse lanes, NBin has 16 lanes feeding the 16 neuron lanes, and NBout has 16 lanes. Each neuron lane is connected to 16 synapse lanes, one from each of the 16 filter lanes. The unit has 256 multipliers and 16 17-input adder trees (16 products plus the partial sum from NBout). The number of neuron lanes and filters  per unit are design time parameters that could be changed. All lanes operate in lock-step.\\ \BASE is designed with the intention to minimize off-chip bandwidth and to maximize on-chip compute utilization. The total per cycle weight bandwidth required by all 16 units of a node is 4K weights per cycle, or 8TB/sec assuming a 1GHz clock and 16-bit weights. The total SB capacity is designed to be sufficient to store all weights for the layer being processed (32MB or 2MB per unit) thus avoiding fetching weights from off-chip. Up to 256 filters can be processed in parallel, 16 per unit. All inter-layer activation outputs except for the initial input and final output are also stored in an appropriately sized central eDRAM, or \textit{Neuron Memory} (NM). NM is shared among all 16 units and is 4MB for the original design. The only traffic seen externally is for the initial input, for loading the weights once per layer, and for writing the final output.

Processing starts by reading from external memory: 1) the filter weights, and 2) the initial input. The filter weights are distributed accordingly to the SBs whereas the activation input is fed to the NBins. The layer outputs are stored through NBout to NM and then fed to the NBins for processing the next layer. Loading the next set of weights from external memory can be overlapped with the processing of the current layer as necessary. Multiple nodes can be used to process larger DNNs that do not fit in the NM and SBs available in a single node. NM and the SBs are implemented using eDRAM as the higher the capacity the larger the inputs and filters that can be processed by a single chip without forcing external memory spilling and excessive off-chip accesses.  

\subsection{\ZFL}

Figure~\ref{fig:nodes-overview}(b) shows a \ZF unit that offers the same computation bandwidth as a \BASE unit. The front-end comprising the neuron lanes and the corresponding synapse lanes is partitioned into 16 independently operating subunits, each containing a single neuron lane and 16 synapse lanes. Each synapse lane processes a different filter for a total of 16.

In the original \ZF design, activations are stored in a ``\textit{Zero-Free Neuron Array Format}'', or ZFNAf. For the time being suffices to know that each activation now is encoded as a $(activation,offset)$ pair in memory where the \textit{activation} field is the raw value, and the \textit{offset }field encodes the coordinates of the activation. Every cycle, each subunit fetches a single $(activation,\mathit{offset})$ pair from NBin, uses the offset to index the corresponding entry from its SBin to fetch 16 weights and produces 16 products, one per filter. The backend is unchanged. It accepts the $16\times 16$ products from 16 subunits which are reduced using 16 adder trees. The adder trees produce 16 partial output activations which the unit accumulates using 64 NBout entries.  The subunit  NBin is 64 entries deep with each entry containing a 16-bit fixed-point value plus an offset field. The total SB capacity remains at 2MB per unit as per the original \BASE design, with each subunit having an SB of 128KB. Each subunit SB entry contains $16\times 16$~bits corresponding to 16 weights. In summary, each subunit corresponds to a single neuron lane and processes 16 weights,  one per filter. Collectively, all subunits have 16 neuron lanes, 256 synapse lanes and produce 16 partial output activations each from a different filter.

\begin{figure}[t]
\centering
\includegraphics[scale=0.45]{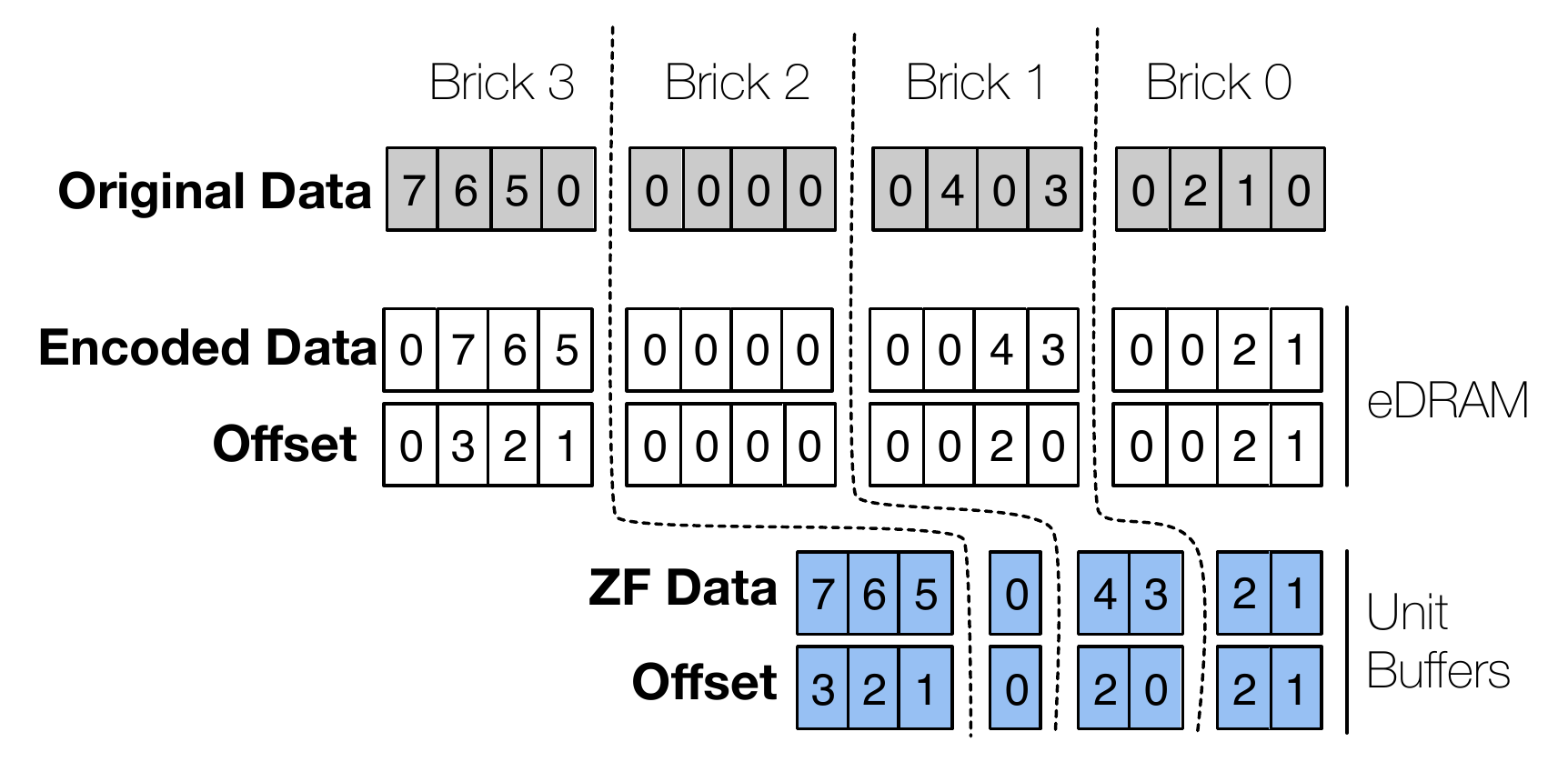}
\caption{Top: \ZFNAf for 4-element bricks. \ZF uses 16-element bricks. Bottom: NBin store format. From Albericio  \textit{et al.}~\cite{cnvlutin}}.
\label{fig:encoding}
\end{figure}

Figure~\ref{fig:encoding} shows the Zero-Free Neuron Array format (\ZFNAf)  that enables \ZF to avoid computations with ineffectual activations. in ZFNAf only the effectual activations are stored, each along with an offset indicating its original position. The ZFNAf is generated at the output of the preceding layer, where it typically would take several tens of cycles or more to produce each activation.  
\ZFNAf encodes activations as $(value, \mathit{offset})$ pairs in groups, or \textit{bricks}. Each brick corresponds to a fetch block of the \BASE design, that is an aligned, continuous along the input features dimension $i$ group of 16 activations, i.e., they all have the same x and y coordinates. Bricks are stored starting at the position their first neuron would have been stored in the conventional 3D array format adjusted to account for the offset fields and are zero padded. Thr grouping in bricks maintains the ability to index the activation array in the granularity necessary to process each layer.

\subsubsection{Skipping the Ineffectual Activations}

 \BASE fetches a single fetch block of 16 activations per cycle which it broadcasts to all 16 units. This blocks contains work for all synapse lanes across 256 filters. In order to keep the neuron lanes busy as much as possible, \ZF assigns work differently to the various neuron lanes. Specifically, while \BASE, as originally described, used an \textit{activation interleaved} assignment of input activations to neuron lanes (i.e., if neuron lane 0 was given activation $a(x,y,i)$, then neuron lane one would be given $a(x,y,i+1)$), \ZF uses a \textit{brick interleaved} assignment (which is compatible with \BASE as well).  For example, if neuron lane is processing an activation brick starting at $a(x,y,i)$, neuron lane 1 would be given the brick starting at $a(x,y,i+16)$. Specifically, \ZF  divides the window evenly into 16 \textit{slices}, one per neuron lane. Each slice corresponds to a complete vertical chunk of the window (all bricks having the same starting $z$ coordinate). Each cycle, one activation per slice is fetched resulting into a group of 16 activations, one per lane, thus keeping all lanes busy. For example, let $e(x,y,z)$ be the $(activation, \mathit{offset})$ pair stored at location $(x,y,z)$ of an input array in \ZFNAf. In cycle 0, the encoded activations at position $e(0,0,0)$, $e(0,0,16)$, ..., $e(0,0,240)$ will be fetched and broadcast to all units and processed by neuron lanes 0 through 15, respectively. As long as all 16 bricks have a second effectual activation, in cycle 1, $e(0,0,1)$, $e(0,0,17)$, ..., $e(0,0,241)$ will be processed. If, for example, brick 0 had only one non-zero neuron, in the next cycle the first encoded activation that will be fetched will be $e(1,0,0)$ assuming an input activation depth $i$ of 256. 

\subsubsection{Fetching the Effectual Activations}

To avoid performing 16 independent, single-activation-wide NM accesses per cycle, \ZF uses a \textit{dispatcher} unit that makes 16-activation wide accesses to NM while keeping all neuron lanes busy. For this purpose, the subarrays the NM is naturally composed of are grouped into 16 independent banks and the input neuron slices are statically distributed one per bank. While the dispatcher is physically distributed across the NM banks.

\begin{figure}[]
\centering
\includegraphics[scale=0.6]{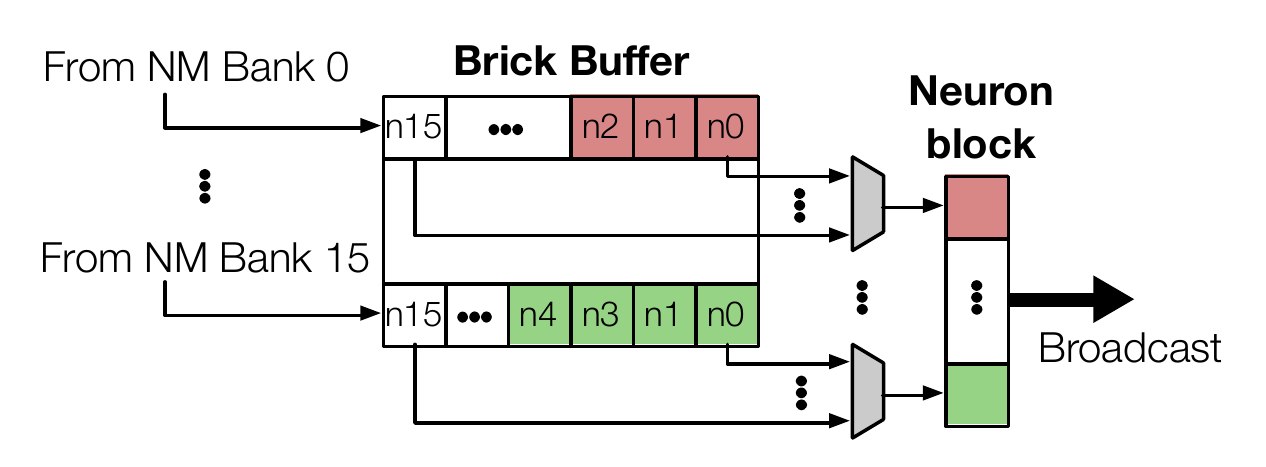}
\caption{Dispatcher~\cite{cnvlutin}}
\label{fig:dispatcher}
\end{figure}

Figure~\ref{fig:dispatcher} shows that the dispatcher has a 16-entry Brick Buffer ($BB$) where each entry can hold a single brick. Each BB entry is connected to one NM bank via a 16-activation-wide bus and feeds one of the neuron lanes across all units via a single-activation-wide connection. Initially, the dispatcher reads in parallel one brick from each bank for a total of 16 activation bricks. In subsequent cycles, the dispatcher broadcasts the non-zero activations, a single activation from each BB entry at a time, for a total of 16 activation, one per BB entry and thus per neuron lane each cycle.  
Before all the effectual activations of a brick have been sent to the units, the dispatcher fetches the next brick from the corresponding NM bank. To avoid stalling for the NM's response, the fetching of the next in processing order brick per bank can be initiated as early as desired since the starting address of each brick and the processing order are known in advance. Since the rate at which each BB will drain will vary depending on the number of effectual activations per brick, the dispatcher maintains a per NM bank fetch pointer.

\subsubsection{Generating the ZFNAf}
The initial input to the CNNs studied are images which are processed using a conventional 3D array format. The first layer treats them as a 3-feature deep neuron array with each color plane being a feature. All other convolutional layers use the \ZFNAf which \ZF generates on-the-fly at the output of the immediately preceding layer. %

In \ZF before writing to the NM, each 16 element output activation group is encoded into a brick in \ZFNAf.  This is done by the \textit{Encoder} subunit. One encoder subunit exists per \ZF unit.
The same interconnect as in \BASE is used, but widened to accommodate the offset fields.
The encoder can afford to do the encoding serially since: 1) output activations are produced at a much slower rate, and 2) the encoded brick is needed for the next layer.

\section{Alternate Encoding of Effectual Activations} 
Earlier we described the ZFNAf format which encodes the effectual neuron values by packing them at the beggining of the brick container. Their offsets were encoded separately using 4 bits per value for a brick of 16 values. This represents a 25\% overhead for 16-bit values and bricks of 16 elements. This section presents alternate activation array formats that reduce memory overhead. For clarity, the discussion that follows uses examples where only zero-value activations are considered as ineffectual. However, the criterion can be more relaxed in practice.

\subsection{RAW or Encoded Format (RoE)}
 Another encoding uses just one extra bit per brick container at the expense of not being able to encode all possible combinations of ineffectual values. Specifically, the first bit of the brick specifies whether the brick is encoded or not. When the brick is encoded the remaining bits are used to store the neuron values and their offsets. As long as the number of effectual activations is such so that they fit in the brick container the brick can be encoded. Otherwise, all activation values are stored as-is and we lose the ability to skip the ineffectual activations for the specific brick. For example, let us assume that we have bricks of size 4 and 16 bit values. In total, each such brick requires $4\times 16=64$ bits. A brick containing the values $(1,2,0,0)$ can be encoded using 65 bits as follows: $(1,(0,1),(1,2))$. The first $1$ means that the brick is encoded. The $(offset,value)=(0,1)$ that follows uses two bits for the offset and 16 bits for the value. In total, the aforementioned brick requires $1+2\times(16+4)=41$ bits can fit within the 65 bits available. A brick containing the values $(2,1,3,4)$ cannot fit within 65 bits and thus will be stored in raw format: $(0,2,1,3,4)$ using 65 bits where the first $1$ is a single bit indicating that the rest of the brick is not encoded and every value is 16 bits long.

\subsection{Vector Ineffectual Activation Identifier Format (VIAI)}

An alternate encoding would leave the activation values in place and  use an extra 16-bit bit vector $I$ to encode which ones are ineffectual and thus can be skipped. For example, assuming bricks of 4 elements a brick containing $(1, 2, 0, 4)$ could be encoded as-is plus a 4 bit  $I$ vector containing $(1101)$. For bricks of 16 activations each of 16 bits, this format imposes an overhead of 16/256, or 6.25\%.

\subsection{Storing Only the Effectual Activations}
 Another format builds on VIAI by storing only the effectual values. For example, a 4 element activation brick of (1,0,0,4) in VIAI would be stored as (1001,1,0,0,4). In the CompressedVIAI it would stored instead as (1001,1,4). Here the two ineffectual zero activations were not stored in memory. Since now bricks no longer have a fixed size, a level of indirection is necessary to support fetching of arbitrary bricks. If the original activation array dimensions are $(X.Y,I)$ then this indirection array $IR$  would have $(X,Y,\lceil I/16\rceil)$ pointers. These can be generated at the output of the preceding layer. 
 
 Further reduction in memory storage can be possible by storing activations at a reduced precision. For example, using the method of Judd \textit{et al.}~\cite{judd:reduced} it is possible to determine precisions per layer in advance based on profiling. 
 It may be possible to adjust precisions at a finer granularity. However, both the pointers and the precision specifier are overheads which reduce the footprint reduction possible.
 
\section{Zero-Memory-Overhead Ineffectual Activation Skipping}

In the original \ZF  implementation the ineffectual activations were ``removed'' at the output of the preceding layer. The ZFNAf incurs a memory storage overhead and the writes and reads of the activation offset values, require additional energy. This section describes an alternate dispatcher design that ``eliminates'' ineffectual activations while fetching them from the NM and prior to communicating these activation values to the tiles.

Specifically, processing for a layer starts by having the dispatcher, as described previously, fetch 16 activation bricks, one brick per neuron lane. The dispatcher then calculates the I (as described previously in the VIAI format) vectors on-the-spot using 16 comparators per brick, one per activation value. The dispatcher then proceeds to communicate the effectual activations at a rate of  one per cycle. When communicating an activation value, the dispatcher will send also the offset of the activation within its containing brick. For example, if the input activation brick contains $(1,0,0,4)$, the dipatcher over two cycles will send to the tiles first $(00b,1)$ ($(offset, value)$) followed by $(11b,4)$. Once all effectual activation values have been communicated to the titles, the dispatcher can then proceed to process another brick for the specific neuron lane. Many options exist for what should be the criterion for detecting ineffectual activations. For example, we could use a simple comparison with zero, a comparison with an arbitrary threshold, or a comparison with a threshold that is a power of two.

\begin{figure}
        \centering
        \includegraphics[width=\columnwidth]{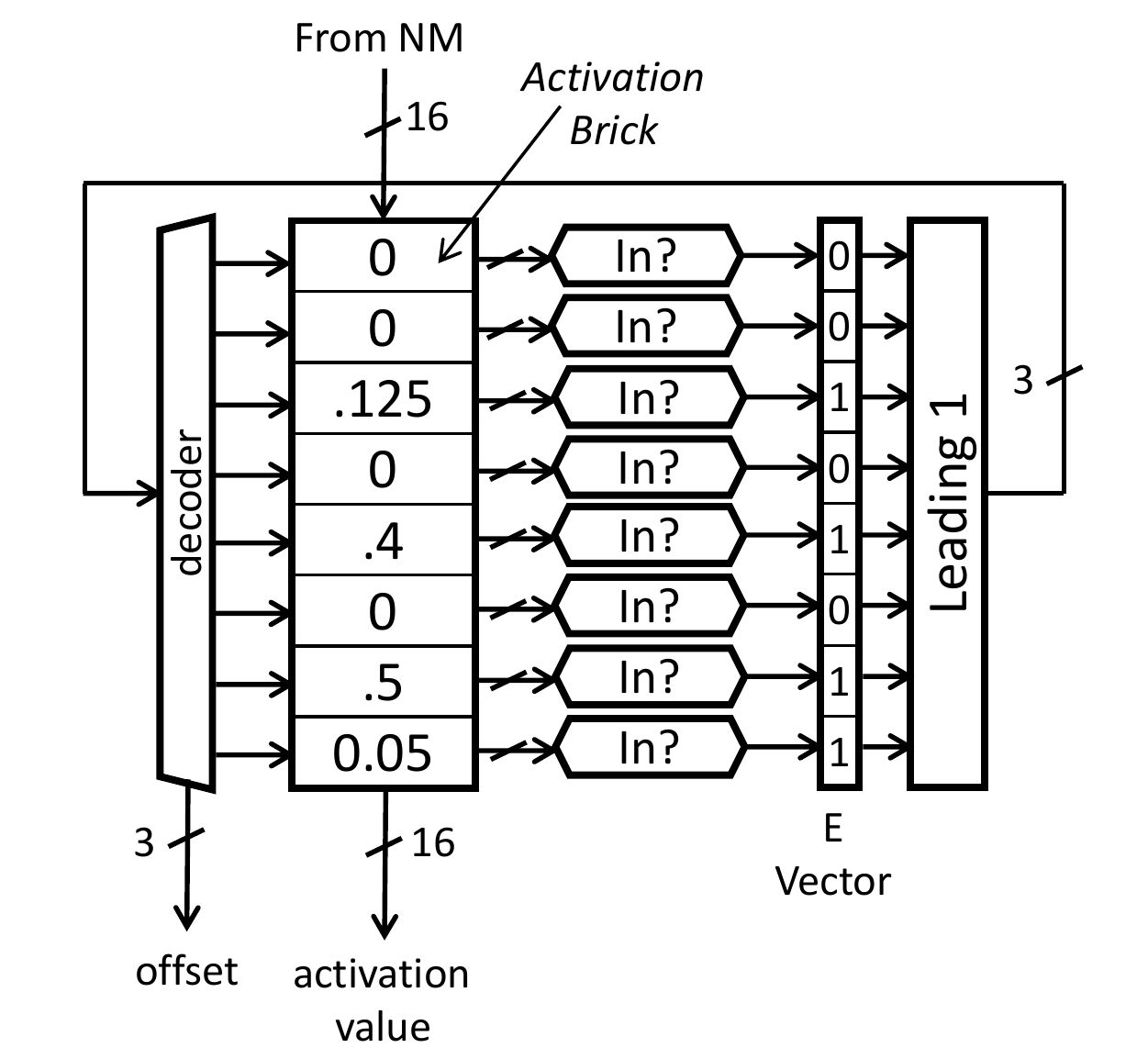}
   
\caption{Detecting and Skipping Ineffectual Activations at the Brick Buffer in the Dispatcher.}
\label{fig:activationskipping}
\end{figure}

Figure~\ref{fig:activationskipping} shows an example, detailed brick buffer implementation of activation skipping in the dispatcher. For clarity, the figure shows only one of the 16 brick buffers and assumes that bricks contain only 8 activations. A second brick buffer per activation lane (not shown) could overlap the detection and communication of the effectual activations from the current brick, with the fetching of the next brick. More such brick buffers may be needed to completely hide the latency of NM.

The figure shows an activation brick that has just been placed into the BB. Next to each BB entry there is an ``ineffectual activation'' (shown as a hexagon labeled as ``In?'') detector. These detectors identify those activations that are ineffectual. As drawn, the output is set to zero if the activation is ineffectual. The collective outputs of these detector form an $E$ vector which drives a ``leading  bit that is 1'' detector. The output of this detector is the offset of the first effectual activation which drives a decoder that reads the activation value out from the BB. The activation value and its offset is then broadcast to the tiles. The E vector position for this activation is reset and the process continues with the next effectual activation. For our example, four cycles would be needed to communicate the four effectual activation values.

\section{Skipping Ineffectual Synapses (Weights)}
\label{sec:weight_skipping}
This section describes \ZFLn which extends \ZFL to also skip ineffectual weights. We have known that a large fraction of weights are ineffectual. For example, once precisions are trimmed per layer as per the methodology of Judd \textit{et al.}~\cite{judd:reduced} a large fraction of weights becomes zero. Most likely, additional weights are ineffectual, for example, weights whose value is near zero. Other work has shown that networks can be also be trained to increase the fraction of weights that are ineffectual~\cite{weightsharing}. Different than activations, weight values are available in advance and thus identifying which are ineffectual can be done statically. This information can be encoded in advance and conveyed to the hardware which can then skip the corresponding multiplications at run-time even when the corresponding activation value is non-zero (or, in general, effectual depending on the criterion being used for classifying activations as ineffectual). 

As described, each cycle, \ZFL processes 16 activations in parallel across 16 filters per unit. The number of activations and filters per unit are design parameters which can be adjusted accordingly. For the purposes of this discussion we will describe \ZFLn assuming that both are 16.

Without loss of generality let us assume that the input neuron array has a depth of 256 and that the window stride is 1. For clarity, let us use $n^B(x,y,i)$ to denote an activation brick that contains $n(x,y,i)...n(x,y,i+15)$ and where $(i\   MOD\  16) = 0$. Similarly, let $s^Bf(x,y,i)$ denote a weight brick containing weights $s^f(x,y,i)...s^f(x,y,i+15)$ of filter $f$ and where again $(i\  MOD\  16)=0$.

For the purposes of this discussion we assume that for each input activation brick $n^B(x,y,i)$, a 16-bit vector $I^B(x,y,i)$ is available, whose bit $j$ indicates whether activation $n^B(x,y,i+j)$ is ineffectual. There is one $I^B(x,y,i)$ vector per input activation brick, hence $i$ is divisible by 16. As with ZFNAf, the $I$ vectors can be calculated at the output of the previous layer, or at runtime, as activation bricks are read from NM as per the discussion of the preceding section. We also assume that for each weight brick, similar $IS$ vectors are available.  Specifically, for each weight brick $s^Bf(x,y,i)$ where $f$ a filter, there is a 16-bit bit vector $IS^B_f(x,y,i)$ which indicates which weights are ineffectual. For example, bit $j$ of $IS^B_0(x,y,i)$ indicates whether weight $s^0(x,y,i+j)$ (filter 0) is ineffectual. The $IS$ vectors can be pre-calculated and stored in an extension of the SB.

Without loss of generality, let us assume that at some cycle $C$, \ZFL starts processing the following set of 16 activation bricks in its 16 neuron lanes: Neuron lane 0 would be processing activations $n^B(x,y,0)$ while neuron lane 15 would be processing $n^B(x,y,240)$. If all activation values are effectual 16 cycles would be needed to process these 16 activation bricks. However, in \ZFL the activation bricks will be encoded so that only the effectual activations are processed. 

In that case, all neuron lanes will wait for the one with the most effectual activations before proceeding with the next set of bricks. Equivalently, the same is possible if the positions of the effectual activations per brick are encoded using the aforementioned $I$ vectors. The dispatcher performs a leading zero detection on the $I$ vector per neuron lane to identify which is the next effectual activation to process for the lane. It then proceeds with the next zero bit in $I$ until all effectual activations have been processed for the lane. When all neuron lanes have processed their effectual activations, all proceed with the next set of bricks.

Since now the $IS$ vectors are also available at the dispatcher needs to do is to take them into account to determine whether an activation ought to be communicated. Specifically, since each activation is combined with 16 weights, each from a different filter, an effectual activation could be skipped if \textit{all} corresponding weights are ineffectual. That is, each neuron lane can combine its single $I$ vector with the 16 $IS$ vectors for the corresponding weight bricks to determine which activations it should process. Specifically, a neuron lane processing $n^B(x,y,i)$ calculates each bit $j$ of a $Can\ Skip$ 16-bit vector as follows:

\begin{equation}
\label{eq:CanSkip}
Can\ Skip^B(x,y,i,j) = \prod_{f=0}^{15}IS^B_f(x,y,j) + I^B(x,y,j)
\end{equation}

and where the operations are boolean: the product is an AND and summation is an OR.
That is, an activation value can be skipped if the activation is ineffectual as specified by  $I$ (activation vector) or if \textit{all} corresponding weights are ineffectual. The higher the number of filters that are being processed concurrently, the lower the probability that an otherwise effectual activation will be skipped. For the original DaDianNao configuration which uses 16 tiles of 16 filters each, 256 weights, one per filter, will have to be ineffectual for the activation to be skipped. However, pruning has been known to be able to identify ineffectual weights and retraining has been known to increase the number of ineffectual weights. Both will increase opportunities for skipping additional neurons beyond what is possible with \ZFL.
Moreover, other configurations may process fewer filters concurrently, thus having a larger probability of combining an activation with weights that are all ineffectual.

It can be observed that in Equation~\ref{eq:CanSkip} all the $IS$ product terms  are constants. As described in \ZFL the same set of 16 weight bricks will be processed concurrently over different windows. Accordingly, the $IS$ products (first term of the sum) can be precalculated and only the final result need to be stored and communicated to hardware. For a brick size of 16 and for tiles that process 16 filters concurrently, the overhead drops from 16 bits per brick to 16 bits per 16 bricks. Assuming 16-bit weights, the overhead drops from $\frac{1}{16}^{th}$ to $\frac{1}{256}^{th}$.

\begin{figure*}
        \centering
        \includegraphics[width=\textwidth]{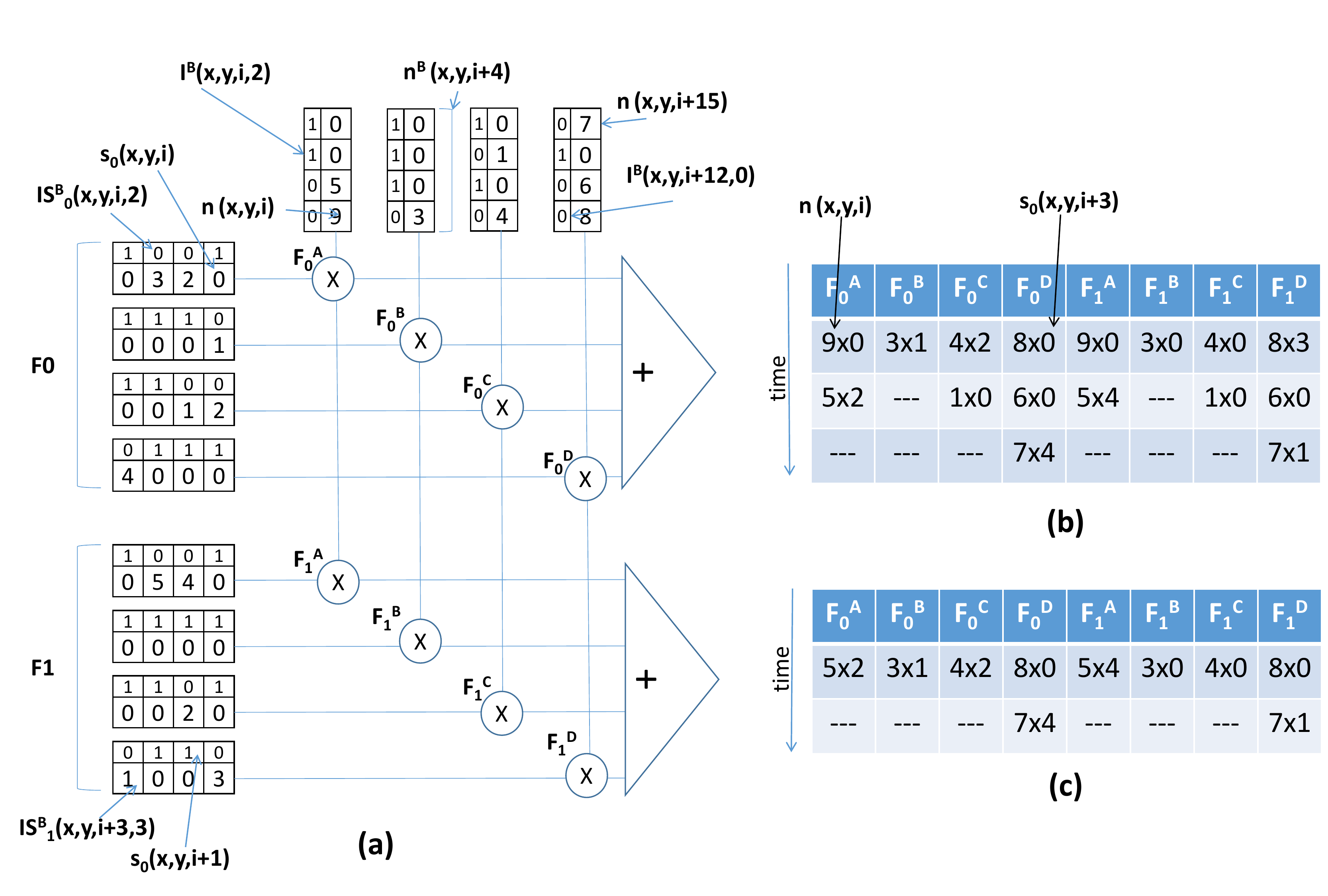}
   
\caption{\ZFLn:An example showing the skipping of weights and activations. (a) Processing bricks of 4 elements each on a tile that processes 2 filters and 4 weights per filter. (b) Execution progression in \ZFL. (c) Execution in \ZFLn.}
\label{fig:weightskipping}
\end{figure*}

\noindent\textbf{\ZFLn -- An Example:}
Figure~\ref{fig:weightskipping} 
shows an example of how \ZFLn operates. For clarity, the example assumes that the brick size is 4 and shows a tile that processes two filters in parallel and two weights (synapses) per filter. As part (b) shows it takes 3 cycles to process all input bricks as activation (neuron) brick $n^B(x,y,i+12)$ contains 3 effectual activations. However, as part (c) shows, one of these effectual activations, specifically, $n(x,y,13)=6$ would have been combined with weights $s^0(x,y,13)$ and $s^1(x,y,13)$ which are both 0 and hence ineffectual. \ZFLn skips this computation and now the input activation bricks can all be processed in just 2 cycles. Additional effectual activations are skipped as well as they would have been combined with ineffectual weights.

\section{Conclusion}
\label{sec:theend}

We presented a set of modification and extension to the \ZF CNN accelerator. The modifications change the memory storage and energy tradeoff by encoding ineffectual activations differently than the original \ZF proposal. We also presented a technique that identifies the ineffectual activations while reading them from memory imposing no memory storage and access overheads. Finally, we decribed an extension to \ZF that can benefit from ineffectual weights also.
\section*{Acknowledgments} This work was supported by an NSERC Discovery Grant.

\Urlmuskip=0mu plus 1mu\relax
\bibliographystyle{ieeetr}
\bibliography{ref}

\end{document}